# The role of successful human-robot interaction on trust

## Findings of an experiment with an autonomous cooperative robot

Nadine Bender
Corporate Research
KUKA
Augsburg, Germany
Nadine.Bender@kuka.com

Samir El Faramawy
Corporate Research
KUKA
Augsburg, Germany
Samir.ElFaramawy@kuka.com

Johannes Maria Kraus
Human Factors
Ulm University
Ulm, Germany
johannes.kraus@uni-ulm.de

Martin Baumann
Human Factors
Ulm University
Ulm, Germany
martin.baumann@uni-ulm.de

## ABSTRACT

The foundation of this paper is an experiment of 15 participants interacting directly with an autonomous robot. The task for the participants was to carry a table, in two different setups, together with a robot, which is intended to support older people with heavy lifting tasks. By collecting and analyzing observational, quantitative, and qualitative data the interaction was investigated with a specific emphasis on trust in the robot. The overall aim was a better understanding of people's emotional and evaluative reactions when they engage with a functioning robot in a relatable everyday scenario. This study shows that successful cooperative task completion has a positive effect on trust and other related evaluations, like the perceived adaptiveness regarding the robot's behavior.

## 1 Introduction

Since the very beginning of automation, the clear purpose of robots has been to help humans. For the first 40 to 50 years of applied robotics, their use was mainly restricted to manufacturing processes. There they relieved workers mostly of physically straining tasks and hence improved the physical well-being of humans in the working age. With the turn of the centuries, both research and technology development have been advancing into additional application fields. Robots are already spreading from factory buildings to hospitals, public spaces, and personal homes. Still, the prominent goal is to improve conditions for humans, however with changing environments, tasks, and users. This also means that it becomes more important to consider the role of technology acceptance [9, 16, 17] and users' trust [22, 26] in robots to increase the likelihood of successful implementation in these new application areas. Only by understanding the processes leading to trust in robots, future technology development can be human-centered, enhancing safe and efficient adoption of robots. Kok and Soh support the importance of trust in human-centered technology development: "Without appropriate trust, robots are vulnerable to either disuse or misuse. In this respect, trust is an enabler that allows robots to emerge from their industrial shell and out into the human social environment" [33, p. 305]. After all, the biggest challenge in designing social robots is to make them appealing to people so that they accept the robots in their private life [2].

In this study, a robot demonstrator developed in the project RoSylerNT, funded by the German Federal Ministry of Education and Research (BMBF) was evaluated. The investigated robot Rosy is a Socially Assistive Robot (SAR) which is a mixture of socially engaging robots, using symbols, gestures or speech, and assistive robots who assist physically or neurologically disabled people [1]. Rosy is envisioned to assist older people at home for example with carrying tasks. The goal of this study was to investigate realistic and direct human-robot-interaction (HRI) with an autonomously acting robot. This contrasts with most HRI studies, where participants either have to imagine the interaction, are shown pictures or videos or in the best case get to interact in a Wizard of Oz experiment. Also, the antecedents of trust in realistic HRI were investigated along some of the variables of the Almere model [9]. The main research questions were how the experience to succeed vs. to fail to successfully cooperate with a robot influences trust in the robot.

## 2 Related Work

Based on the Almere model, this research investigates robot evaluation in a realistic study setting with an autonomous robot with a specific focus on trust and its antecedents. Thus, it builds mainly on three realms of research: The realism of HRI studies, technology acceptance modeling, and trust in automation.

### 2.1 Realism of HRI studies

In most HRI studies, participants are shown pictures or videos to immerse virtually in the interaction. This however is very different from directly interacting with a real robot. Fazio and Zanna in this regard state that, "support has been found for the notion that attitudes based on direct, behavioral experience with an attitude object are more predictive of later behavior than are attitudes based on indirect, nonbehavioral experience" [14, p. 172]. Research also shows that attitudes based on direct experience are more extreme and less ambivalent [15]. Following that, several authors report that actual interaction with robots is associated with improved attitudes towards robots over time [8, 13]. In this study, participants experienced an untarnished, direct physical interaction with an autonomous robot. Thereby, this study provides a possibility to investigate the relation between HRI experiences and the associated subjective reactions as well as their influence on future behavior and use. The increased realism is important for robot manufacturers as they can receive important feedback on how to design their robots to raise future sales.

## 2.2 Cherry-picking from the Almere Model

To analyze HRI, different models of technology acceptance have been introduced. The history of research sprung from the Technology Acceptance Model (TAM) [16] and has ever since been probed and evolved, for example into the Unified Theory of Acceptance and Use of Technology (UTAUT) model [17]. These models have been adapted to the research field of HRI by Heerink et al. [9], explaining the intention to use with a broad array of belief-, attitude- and disposition-based variables.

| Code | Construct | Definition |
|---|---|---|
| ANX | Anxiety | "subjective, consciously perceived feelings of tension and apprehension" [30] |
| PAD | Perceived adaptiveness | The perceived ability of the system to adapt to the needs of the user |
| PENJ | Perceived Enjoyment | Feelings of joy/pleasure associated with the use of the system |
| PEOU | Perceived Ease of Use | The degree to which one believes that using the system would be free of effort |
| PS | Perceived Sociability | The perceived ability of the system to perform sociable behavior |
| PU | Perceived Usefulness | The degree to which a person believes that the system would be assistive |

**Table 1: Overview of constructs derived from Almere's 13 constructs**

As for the underlying goal of this research to provide initial findings on which subjective antecedents relate to trust differences, in this study, several constructs from the Almere Model were investigated as a potential foundation for the evaluation. The applicability of several of the constructs included in the Almere Model to examine the acceptance of assistive social agents by older people was shown in the research by [9].

To investigate the role of trust antecedents, only the depicted constructs of the Almere Model in table 1 were investigated resembling affective reactions and perceived robot characteristics, and features of the interaction with the robot.

## 2.3 Trust

With the spread of robots from industrial use cases into private homes and health care settings, the requirements of for example user-friendliness towards the robot increase. It is no longer enough to ensure technical excellence and the highest safety and accuracy standards. Therefore, trust is crucial for a successful and safe introduction of assistive robots. "Especially with regard to critical decisions, trust plays an important role in human interactions and could therefore help to increase the robot's acceptance in its role as a collaborative partner" [31, p. 585]. In both interpersonal as well as human-machine relationships it has been found that "the presence, growth, erosion, and extinction of trust have powerful and lasting effects on how each member of any shared relationship behaves and will behave in the future" [20, p. 522].

Trust in automation (TRU) was defined as "the attitude that an agent will help achieve an individual's goal in a situation characterized by uncertainty and vulnerability" [27, p. 51] and has been investigated intensively for 30 years in human-machine interaction. Besides application areas like automated driving [24] and plant simulations [27], recently trust in robots has been broadly discussed and investigated as a prerequisite for human-robot interaction [28]. As research showed, in healthcare robotics the intention to use declines significantly without the users' trust in robots [18]. The important role of trust in HRI is nicely illustrated by Olaronke et al., who state "trust facilitates the reliance of human beings on the ability of social robots to perform their tasks" [19, p.52]. Learned trust in automation has been broadly discussed as a dynamic attitude that is built up in specific, in the early interaction (or relationship building) with new technology, in which available information is used to calibrate one's trust [29].

Considering this research's interest in the role of interaction failures on trust in robots, Desai et al.'s work on the primacy-recency bias is relevant. They found that the earlier the robot's lower reliability in the interaction occurs, the more detrimental the impact on trust is [21]. However, research also shows that robot's warnings ahead of potential performance drops were able to prevent declines in the user's trust ratings [22]. Based on these findings, it can be assumed, that with more communication and help on the side of the robot, such failures might be prevented. This line of argumentation can be supported by Kraus et al. [23] who pose that trust takes time to build and is at least temporarily diminished after automation failures without any explanations to the cause. Taken together, performance, reliability, and communication seem to essentially influence trust levels. Also in autonomous driving, Choi et al. already manifested in their paper that, technical competence influences trust [32]. Interestingly, the construct of perceived adaptiveness, used in this paper, consists of very similar items as Choi et al.'s construct technical competence. This implies convergence between the two constructs to a certain degree. Our results will show that the findings from studies in autonomous driving may also be valid for the interaction with robots.

## 2.4 Hypotheses

Against this background, the following two hypotheses are generated and lead our further investigations.

H1: *People who experience successful interaction with a robot, trust the robot more (TRU) and show corresponding effects in potential trust antecedents (ANX, PAD, PENJ, PEOU, PS, PU).*

H2: *Successful interaction induces higher levels of trust (and associated evaluations) while unsuccessful interaction on the contrary leads to corresponding reductions.*

## 3 Method

The study took place on the company premises in the robot lab of KUKA's Corporate Research department. For a week each day, three participants were invited to interact with the robot. According to Covid-19 health regulations, both operators and participants had to wear FFP2 face masks and all surfaces were disinfected after each participant. Participants were also not allowed to be on the premises at the same time. Ethical approval of the study was obtained by the Ethical Committee of the German Sport University Cologne (DSHS) and KUKA's worker's council.

In total seven male and eight female participants in the age of 18-49 years were recruited using the following criteria: Age between 18 and 49, healthy and fit, training several times a week, no known pregnancy, normal vision and hearing abilities, not in a high-risk corona group. The decision to exclude older people was met because the demonstrator has not yet been cleared with a CE marking and due to the Covid-19 pandemic, the risk to include older people would have been too high.

### 3.1 Robot description

The robot setup consisted of a 3-DoF KUKA mobile platform, two 6-DoF KUKA LBRiiwa robots, a Pan-Tilt-Unit with a Roboception camera, and an iPad mounted between both arms. During the project, several features were developed and integrated:

- 3D-sensors, as well as lasers and radar, were added for increased perception: the robot can not only navigate using a map but also see and react to obstacles and humans, without the interference of the operator
- Control method which allows controlling of the complex kinematic chain
- Gesture and face recognition
- Interaction concept for visual feedback via a tablet
- Theoretical work for human pose recognition to realize an ergonomically optimal position for the user
- Tested, but not included in the demonstrator: Environmental modeling in 3D

The testing area was a 4m x 9m areal which was separated into two "rooms" by a concrete column and a low fence. To reach the other room, the participants had to pass a space of 1.9m to clear the column. The total length of the robot and table was approximately 2.25m with a width of 70cm.

### 3.2 Procedure

After arriving at the site, participants were informed about potential study risks, their rights and agreed to the study procedures and data protection policy. Before being led to the testing area and hence before actually seeing any robot, they were given a first short questionnaire. Next, they were equipped with EEG caps and markers for a research question not in the scope of this study. During this process, the participants were explained their task.

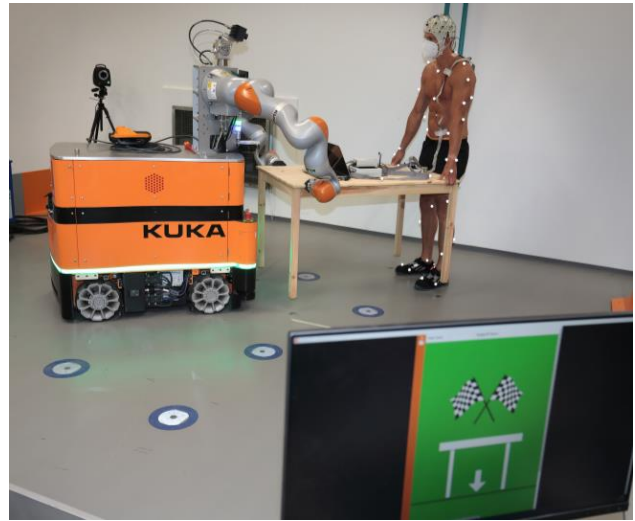

**Figure 1: Participant carrying the table in cooperation with the robot. The monitor shows information that is also shown on the iPad. Participant agreed to the publication of the photo.**

In the first part of the study, participants carried the table with another human. One trial consisted of moving the table from room A to room B, the next consisted of moving the table from room B to room A without changing the direction of the participant. This meant that the table was alternatively pulled and pushed. In total, the participants completed 12 trials (6 pulls and 6 pushes) with the human before repeating the same task another twelve times with the robot. An interaction failure was recorded when the robot stopped because the participant steered the robot too close to an obstacle and the robot had to avoid a collision. Before the interaction, the participants were informed of the robot's three-light set up to communicate: a green light indicates that all is well. The yellow light signals a possible dangerous spatial position of the robot. If the position isn't adjusted, e.g. the robot steered away from the obstacle, the screen turns red and shows a warning sign. In general, it is important to note that the robot was always in the role of the assistant, and the control lay with the human. Both during the start and end of the interaction, as well as in the phase of carrying, the human was in authority.

After the participants completed the study protocol, during which they were observed using a structured observation protocol, the cap and markers were removed, and the participants were led into another room. There they completed more questionnaires and were interviewed, as described in detail in the next paragraph. All in all, they spent between three and four hours on the premises and received a 50 € compensation for their efforts.

### 3.3 Data collection

Most literature on robot evaluations works either with self-created questionnaires [4, 5] or with observations [6] or interviews of users or experts [4, 6, 7, 8]. Since the complex study setting allowed only for a restricted number of participants, triangulation of methods was used and included all of the above-mentioned methods.

In the quantitative part, in the pre-questionnaire socio-demographics and prior experience with robots were assessed. Furthermore, the Negative Attitude towards Robots Scale was applied in a pre-and post-questionnaire before and after the interaction [3]. In the post-questionnaire. the included constructs of the Almere Model (see table 1) [9] were assessed as well as trust in automation with the Checklist for Trust between People and Automation [10].

In addition to the quantitative data, qualitative data was gathered with interviews and observations. In semi-structured interviews, the participants were asked to reflect on the interaction and express themselves freely. This additional information is used to enhance the understanding of user attitudes and acceptance towards the robot [6,7,11]. The interview guide was structured in three parts:

1) Actual interaction: Questions about the first impression, appearance, predictability of robot behavior, communication, etc.
2) Potential use of the robot: Questions about the use in the target group/area of application, potentials, concerns, suggestions for improvement
3) Conclusion with comments and a final evaluation

### 3.4 Method of analysis

Due to the sample size of 15 observations, for group comparisons, non-parametric significance tests were conducted. To analyze the correlations properly, Spearman's rank correlation coefficient is calculated to show monotonic correlation [12]. For the investigation of the effect of interaction success vs. failure, the sample was split in two groups and tested with the Mann-Whitney-U Test [24].

### 4 Results

The data acquired for NARS and its subscales showed insufficient internal consistency and therefore had to be excluded from the analysis. The internal consistency of the Almere constructs however was satisfying. It was assessed with Cronbach's Alpha as depicted in table 2. While in the case of ANX (.61) internal consistency is at least acceptable, it is very good for PENJ (.93) and TRU (.89) [25]. For the trust scale, the negatively framed items lead to a high number of outliers and therefore the trust scale was constructed by four of the positively framed items of the Jian et al. [10] scale.

In general, table 2 shows absolute values indicating positive tendencies for PEOU, PU, PAD, PS, PENJ, and TRU being on average above three. ANX on average was the construct with the lowest and PENJ the construct with the highest agreement.

| Construct (Code) | Number of Items | Cronbach's Alpha | M | SD |
|---|---|---|---|---|
| Anxiety (ANX) | 4 | .61 | 2.13 | .60 |
| Perceived Adaptiveness (PAD) | 3 | .76 | 3.67 | .77 |
| Perceived Enjoyment (PENJ) | 5 | .93 | 3.93 | .88 |
| Perceived Ease of Use (PEOU) | 5 | .76 | 3.72 | .79 |
| Perceived Sociability (PS) | 4 | .71 | 3.28 | .66 |
| Perceived Usefulness (PU) | 3 | .83 | 3.11 | .97 |
| Trust (TRU) | 4 | .89 | 3.65 | .95 |
| Note: M: mean; SD: standard deviation | | | | |

**Table 2: Overview of Cronbach's Alpha and descriptive statistics**

First, the focus lies on differences in the post-interaction experiences of participants who caused a failure, by steering too close to an obstacle or ending the interaction before the robot signaled its readiness with the robot and those who cooperated with the robot successfully. The means for the respective groups are depicted in table 3.

Supporting H1, TRU was higher in the successful interaction group and the difference between the groups was indicated by a significant in the Mann-Whitney-U Test ($U(7,8) = 10$, $z = -2.1$, $p = .04$).

In terms of H2, table 3 shows that PAD is higher in the successful interaction group. The participants in the successful interaction group approve Rosy's ability to adapt to their needs more than the participants with failures in their interactions. This difference is significant ($U(7,8) = 10$, $z = -2.13$, $p = .04$). This observation is backed by a participant's statement who felt in control of the robot:

> *"Well, I had the feeling that I was in control. Yes, so he [the robot] only did what I basically told him to do. If I pushed backward or forwards, then he did that and otherwise he would have stopped."* (P5)

For all remaining Almere constructs no significant group differences were found. Yet, most absolute values were somewhat higher in the no-failure than in the failure group and might be addressed in future studies with larger samples and increased statistical power.

|  | ANX | PAD | PENJ | PEOU | PS | PU | TRU |
|---|---|---|---|---|---|---|---|
| No-failure $n = 8$ | 2.06 | 4.04* | 4.23 | 3.95 | 3.50 | 3.08 | 4.16* |
| Failure $n = 7$ | 2.21 | 3.24* | 3.60 | 3.46 | 3.04 | 3.14 | 3.07* |
| *$p < .05$; Tested with Mann-Whitney - U ||||||||

**Table 3: Comparison between means of constructs divided by failures in HRI**

The relationships between the investigated variables can be inspected in table 4 showing Spearman's rank correlation coefficients. Strong correlations between PEOU and PAD (.7), PEOU and ANX (-.6), as well as PENJ and PS (.7), were found. In the same manner, PAD correlates negatively with ANX (-.7) and positively with TRU (.7).

|  | ANX | PAD | PENJ | PEOU | PS | PU | TRU |
|---|---|---|---|---|---|---|---|
| ANX | 1 | | | | | | |
| PAD | -.74** | 1 | | | | | |
| PENJ | -.12 | .33 | 1 | | | | |
| PEOU | -.63** | .71** | .58* | 1 | | | |
| PS | -.11 | .43 | .72** | .60** | 1 | | |
| PU | -.27 | .44* | .40 | .54* | .54* | 1 | |
| TRU | -.45* | .74** | .45* | .49* | .52* | .32 | 1 |
| **$p < 0.01$; *$p < 0.05$ |||||||||

**Table 4: Correlation matrix**

The mentioned correlations are significant on a 1% level but should not be overestimated as for the small sample size. Taken together, these results point in the direction that the PEOU and PAD are important antecedents of TRU in this setting.

These findings are supported by the participants' statements in their interviews. Regarding the correlation between PEOU and PAD participant 1 stated:

> "In the beginning, I had two attempts that didn't go so well, because you first have to get a feeling for how the robot reacts. But when you know how to handle it, it works pretty well." (P1)

Similarly, one participant connects his anxiety in the interaction with the perceived adaptiveness in the sense that he felt the robot was more difficult to steer:

> "It was more difficult with the robot, but maybe also because I was afraid of breaking something." I: "You were also more careful then?" P: "Yes, a bit more carefully, I didn't want to crash into the column. One is already careful." (P2)

Another participant, P3 found the interaction pleasant and enjoyable once he got to know the robot, depicting the correlation between PENJ and PS:

> "Well, the system is, let's say, foolproof with the colorful pictures with checkmarks and so on. Yes, as I said, you have to get into the groove a bit, so that you can figure out how it reacts to what. And then when you've got that out of the way, you can handle it quite well." (P3)

The correlation between ANX and TRU was significant in the quantitative analysis, several comments supported such a relation as reported for example in [28, 29]:

> "Well, uncertainty in the technology - it is not yet so advanced, I would claim. You don't have it like that in the daily life yet, that's why and that's a new situation, you have to get used to it first. And that's why there was a bit of uncertainty and trust that was simply missing." (P4)

## 6 Discussion and Outlook

To summarize, the perceived adaptiveness and trust are highly correlated and were significantly lower in the failure than in the no-failure group. This underlines the elevating role of these variables in HRI. The constructs singled out are affected by the interaction and in this instance are lower when a failure in the interaction has been experienced.

> "So the safety must be right. In general, the display always showed red, yellow, green. And I just did not know why it was red, why it was not yellow. I don't know what I did wrong, that would be maybe in the future, that you know, oh okay you are too far to the right, etc." (P2)

The study findings and this quote by participant 2 support an important role of transparency and a comparison of expectations and observed robot behavior for trust in robots. Such a mechanism was also supported in Kraus et al. [23] in the way that the availability of explanations of automation behavior can prevent a trust decrease after a system malfunction. In the case of this study, transparency of the processes involved and reasons for the failed cooperation was not provided. Communication of the robot regarding the failure was very low key and the participants did not get any information afterward from the robot. In this regard, the reported findings of lower trust evaluations in participants who experienced a failed cooperation might be prevented by enhancing the robot's transparency and providing explanations for such situations. In this sense, also a clear emphasis on the robot's adaptability might be a promising direction to enhance trust in the robot.

While the strength of this study is an extensive, detailed investigation of direct, real-life interaction with an autonomous robot in a mixed-method approach combining observations, questionnaires, and interviews, this method is at the same time the reason for the comparatively small sample size. Hence our research provides a HRI'21, March 2021, Virtual Conference    Bender et al.first step and the investigated relationships might be further addressed in future studies with larger samples.

Taken together, HRI holds enormous potential for the future despite all the engagement already in place. What shall be most in focus is getting into practice and studying the general and the field-specific properties when people and robots come together. The investigation of trust in robots and its antecedents in terms of beliefs and attributes is an approach that manages to reveal important characteristics in which sense people are willing to have interactions with robots like Rosy as well as in which way robots shall be adjusted for a cooperative future. A participant stated very clearly her concerns, but also positively referred to trust as a resource for successful and subjective positive interaction with robots in situations of uncertainty and risk:

> *"You always have a little concern. Will it really stop now and not drive over it? I don't know, the concerns are clearly there, but I for one would trust the technology."* (P6)

## ACKNOWLEDGMENTS

The research has been conducted during the project RoSylerNT (16SV7851), funded by the German Federal Ministry of Education and Research (BMBF). Special thanks go to the project partners of the German Sport University Cologne (DSHS) who conducted the study together with us.## REFERENCES

[1] Jeonghye Han, Daniela Conti. 2020: The Use of UTAUT and Post Acceptance Models to Investigate the Attitude towards a Telepresence Robot in an Educational Setting. *Robotics* 9, 2 (May 2020), 1-19. DOI:10.3390/robotics9020034

[2] Christoph Bartneck, Tatsuya Nomura, Takayuki Kanda, Tomohiro Suzuki, Kennsuke Kato. 2005: A cross-cultural study on attitudes towards robots. In *Proceedings of the Conference on Human-Computer Interaction International, (HCI 2005)*, July 22-27, 2005, Las Vegas. DOI: 10.13140/RG.2.2.35929.11367

[3] Tatsuya Nomura, Tomohiro Suzuki, Takayuki Kanda, Kensuke Kato. 2006: Measurement of negative attitudes toward robots. *Interaction Studies* 7, 3 (Nov. 2006), 437-454. DOI: 10.1075/is.7.3.14nom

[4] Sandra Bedaf, Patrizia Marti, Farshid Amirabdollahian, Luc de Witte. 2018: A multi-perspective evaluation of a service robot for seniors. The voice of different stakeholders. *Disability and rehabilitation. Assistive technology* 13, 6 (Aug. 2018), 592-599. DOI: 10.1080/17483107.2017.1358300

[5] Yiannis Koumpouros. 2016: A Systematic Review on Existing Measures for the Subjective Assessment of Rehabilitation and Assistive Robot Devices. *Journal of healthcare engineering* (May 2016), 1-10. DOI: 10.1155/2016/1048964

[6] Theo Jacobs, Birgit Graf. 2012: Practical evaluation of service robots for support and routine tasks in an elderly care facility. In *2012 IEEE Workshop on Advanced Robotics and its Social Impacts (ARSO 2012)*, May 21 - 23, 2012, Munich. IEEE, New York, 46-49. DOI: 10.1109/ARSO.2012.6213397

[7] Ulrike Scorna. 2015: Servicerobotik in der Altenpflege. Karsten Weber, Debora Frommeld, Arne Manzeschke und Heiner Fangerau (Ed.): *Technisierung des Alltags. Beitrag für ein gutes Leben?* Franz Steiner Verlag, Stuttgart: (Wissenschaftsforschung, Band 7), 81-96.

[8] Katie Winkle, Praminda Caleb-Solly, Ailie Turton, Paul Bremner. 2018: Social Robots for Engagement in Rehabilitative Therapies: Design Implications from a Study with Therapists. In *Proceedings of the 2018 ACM/IEEE International Conference on Human-Robot Interaction (HRI '18)*, March 5-8, 2018, Chicago, Illinois. ACM Inc., New York, 289–297. DOI: https://doi.org/10.1145/3171221.3171273

[9] Marcel Heerink, Ben Kröse, Vanessa Evers, Bob Wielinga, 2010: Assessing Acceptance of Assistive Social Agent Technology by Older Adults. The Almere Model. *International Journal of Social Robotics* 2, 4 (Dec. 2010), 361-375. DOI: 10.1007/s12369-010-0068-5

[10] Jiun-Yin Jian, Ann M. Bisantz, Colin G. Drury. 2000: Foundations for an Empirically Determined Scale of Trust in Automated Systems. *International Journal of Cognitive Ergonomics* 4, 1 (Mar. 2000), 53-71. DOI: 0.1207/S15327566IJCE0401_04

[11] Philipp Mayring. 2000: Qualitative Content Analysis. 2000: Retrieved February 3, 2021 from https://www.qualitative-research.net/index.php/fqs/article/view/1089/2385

[12] Naresh K. Malhotra. 2007: *Marketing research: an applied orientation*. (6th. ed.). Pearson/Prentice Hall, Upper Saddle River, NJ.

[13] Christoph Bartneck, Tomohiro Suzuki, Takayuki Kanda, Tatsuya Nomura. 2006: The Influence of People's Culture and Prior Experiences with Aibo on their Attitude Towards Robots. *AI & Society – The Journal of Human-Centred Systems* 21, 1-2 (Jan. 2007), 217-230. DOI:10.1007/s00146-006-0052-7

[14] Russell Fazio, Mark Zanna. 1981: Direct Experience And Attitude-Behavior Consistency. *Advances in Experimental Social Psychology* 14, 161-202. DOI: 10.1016/S0065-2601(08)60372-X

[15] James M. Olson, Gregory R. Maio. 2003: Attitudes in Social Behavior. *Handbook of Psychology* Vol. 5, John Wiley & Sons Hoboken, 299-325. DOI: 10.1002/0471264385.WEI0513

[16] Fred D. Davis. 1989: Perceived usefulness, perceived ease of use, and user acceptance of information technology. *MIS Quarterly* 13, 3 (Sep. 1989), 319-340. DOI: 10.2307/249008

[17] Viswanath Venkatesh, Michael G. Morris, Gordon B. Davis, Fred D. Davis. 2003: User Acceptance of Information Technology: Toward a Unified View. *MIS Quarterly* 27, 3 (Sept. 2003), 425-478. DOI: 10.2307/30036540

[18] Ahmad Alaiad, Lina Zhou. 2014: The Determinants of Home Healthcare Robots Adoption: An Empirical Investigation. *International Journal of Medical Informatics* 83, 11 (Nov. 2014), 825-840. DOI: 10.1016/j.ijmedinf.2014.07.003

[19] Iroju Olaronke, Ojerinde Oluwaseun, Ikono Rhoda. 2017: State Of The Art. A Study of Human-Robot Interaction in Healthcare. *IJIEEB* 9, 3 (May 2017), 43-55. DOI: 10.5815/ijieeb.2017.03.06

[20] Peter A. Hancock, Deborah R. Billings, Kristin E. Schaefer, Jessie Y. C. Chen, Ewart J. de Visser, Raja Parasuraman. 2011: A meta-analysis of factors affecting trust in human-robot interaction. *Human factors* 53, 5 (Sept. 2011), 517-527. DOI: 10.1177/0018720811417254

[21] Munjal Desai, Poornima Kaniarasu, Mikhail Medvedev, Aaron Steinfeld, and Holly Yanco. 2013: Impact of robot failures and feedback on real-time trust. In *Proceedings of the 8th ACM/IEEE international conference on Human-robot interaction (HRI '13)*. March 4-6, 2013, Tokyo, Japan. IEEE Press, 251–258.

[22] Allison Langer, Ronit Feingold-Polak, Oliver Mueller, Philipp Kellmeyer, Shelly Levy-Tzedek. 2019: Trust in socially assistive robots. Considerations for use in rehabilitation. *Neuroscience and biobehavioral reviews* 104, (July 2019), 231-239. DOI: 10.1016/j.neubiorev.2019.07.014

[23] Johannes Kraus, David Scholz, Dina Stiegemeier, Martin Baumann. 2019: The More You Know: Trust Dynamics and Calibration in Highly Automated Driving and the Effects of Take-Overs, System Malfunction, and System Transparency. *Human factors: The Journal of Human Factors and Ergonomics Society* 62, 5 (June 2019), 718 - 736. DOI: 10.1177/0018720819853686

[24] Frederick J. Gravetter, Larry B. Wallnau. 2017: *Statistics for the behavioral sciences* (10th ed.). Cengage Learning, Boston, MA.

[25] Neal Schmitt. 1996: Uses and Abuses of Coefficient Alpha. *Psychological Assessment* 8, 4 (Dec. 1996) 350–353. DOI:10.1037/1040-3590.8.4.350

[26] John D. Lee, Katrina A. See. 2004: Trust in automation: designing for appropriate reliance. *Human factors*, 46(1), 50–80. DOI: 10.1518/hfes.46.1.50_30392

[27] John D. Lee, Neville Moray. 1994: Trust, self-confidence, and operators' adaptation to automation. *International Journal of Human-Computer Studies* 40, 1 (Jan. 1994) 153-184. DOI:10.1006/ijhc.1994.1007

[28] Miller, L. Kraus, J., Babel, F., & Baumann, M. (2021, accepted, in print): More than a Feeling -Interrelation of Trust Layers in Human-Robot Interaction and the Role of User Dispositions and State Anxiety. *Frontiers in Psychology*. DOI: 10.3389/fpsyg.2021.592711

[29] Johannes Maria Kraus. 2020*:* Psychological processes in the formation and calibration of trust in automation. *Dissertation*. Universität Ulm. DOI: 10.18725/OPARU-32583

[30] Masami Horikawa, Akihiro Yagi. 2012: The relationships among trait anxiety, state anxiety and the goal performance of penalty shoot-out by university soccer players. *PloS One* 7, 4 (April 2012), e35727. DOI: 10.1371/journal.pone.0035727

[31] Maha Salem, Gabriella Lakatos, Farshid Amirabdollahian, Kerstin Dautenhahn. 2015: Towards Safe and Trustworthy Social Robots: Ethical Challenges and Practical Issues. *7th International Conference on Social Robotics (ICSR)*, October 26-30, 2015, Paris, France. Springer, Cham., 584-593. DOI:10.1007/978-3-319-25554-5_58

[32] Jong Kyu Choi, Yong Gu Ji. 2015: Investigating the importance of trust on adopting an autonomous vehicle. *International Journal of Human-Computer Interaction*, 31, 10 (July 2015), 692-702. DOI: 10.1080/10447318.2015.1070549

[33] Bing Cai Kok, Harold Soh. 2020: Trust in Robots. Challenges and Opportunities. *Current Robotics Report* 1, (Sept. 2020), 297–309. DOI: 10.1007/s43154-020-00029-y
6